# Learning from Profession Knowledge: Application on Knitting

Nada Matta, Oswaldo Castillo Navetty
Charles Delaunay Institute, Tech-CICO Team, University of Technology of Troyes
12 rue Marie Curie, B.P. 2060, 10010, Troyes, France
nada.matta@utt.fr, navetty@hotmail.com

**Abstract:**
Knowledge Management is a global process in companies. It includes all the processes that allow capitalization, sharing and evolution of the Knowledge Capital of the firm, generally recognized as a critical resource of the organization. Several approaches have been defined to capitalize knowledge but few of them study how to learn from this knowledge. We present in this paper an approach that helps to enhance learning from profession knowledge in an organisation. We apply our approach on knitting industry.

## 1 Introduction

Knowledge management is currently defined as a process of identification, formalization, disseminating and use of knowledge in order to promote creativity and innovation in companies. This process (Figure 1. ) takes into account the transformation and the evolution of tacit to explicit knowledge [10] and of individual to collective knowledge [5].

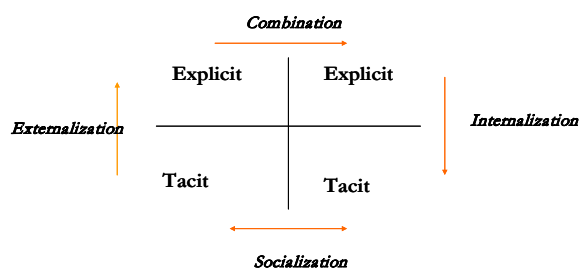

**Figure 1.** Knowledge Management [10].

Methods stemming from the knowledge engineering (such as MASK, REX, KOD, etc.) and the computer supported cooperative work (such as QOC, DIPA, etc.) [6] were used in order to capitalize and make explicit knowledge in an organization. These methods allow defining corporate memories. A corporate memory is defined as the « explicit and persistent representation of the knowledge and the information in an organization » [6], [12]. We can distinguish several types of memories: profession memory, project memory and organization memory.

The sharing and the appropriation of the corporate memories are still real blocking points within organizations. The methods of knowledge management are not sufficient to allow an effective appropriation of the knowledge by the actors of the company. However, the objective of a capitalization of the knowledge is indeed the sharing and the re-use of an experience with the aim of optimizing the process of organizational learning.

The technologies of information and the communication (i.e. semantic web) can constitute a support for sharing. However, Knowledge sharing must be guided if we want to supply the good information at the right moment. Furthermore, the distribution of information is not enough to guarantee the re-use of the knowledge. One condition to reuse knowledge is to be assimilated by the actor that is integrated into his experiences and generate appropriate knowledge at any time in the action [11], [5].

We define an approach based on knowledge management and educational engineering in order to enhance profession knowledge learning in a company. This approach has been applied in textile factory in order to enhance knitting learning. We present in this paper the memory we defined (profession memory in knitting) and the approach we followed to enhance learning in Textile Company.

## 2 Profession memory

We define a profession memory as the externalisation of the knowledge produced in and for a given domain [5]. It represents problem solving. The techniques of knowledge engineering allow the formalization of this type of memory. Several approaches, such as MASK, REX, CommonKADS, or KOD [6], use techniques of knowledge engineering to extract knowledge, to formalize it in conceptual models, where the knowledge that guides problem solving is made explicit. The structure of this type of memory describes generally: the definition of the problem and the process, the problem solving methods, as well as a description of the concepts manipulated in problem solving. This representation of knowledge can lean on graphic completed presentations (for example, the models of problem solving methods of CommonKADS [3] and MASK [7]) and with textual explanations as it can be as problem experience forms indexed with trees of concepts as it is the case of the elements of experience of REX [9].

### 2.1 Using profession memory for learning

According to Gurteen, "sharing is not just about giving. It is about: soliciting feedback, asking questions, telling people what you plan to do before doing it, asking other people for help, asking someone to work with you in some way – however small, telling people what you are doing and more importantly why you are doing it, asking people what they think – asking them for advise, asking people what would they do differently, not just sharing information but know-how and know-why" [8].

Generally, a profession memory is organized to represent know-how in a given domain. It is rather about a practical knowledge acquired from the experience. The descriptions of the context, the problem solving methods as well as the evolution of the activity are not sufficient to establish a complete training on a profession. The learning techniques which we intend to define address essentially actors in a domain, aiming to learn some experience of an expert in this domain.

Let us note that an expertise is represented as conceptual models with models (model of task, activity, domain, etc.). These very useful models for the knowledge extraction are not easily accessible for knowledge appropriation. It is then important to reorganize theses models to answer the various needs of the organization actors.
The CSAO approach we defined propose to reorganise models defined in a profession memory in order to facilitate their appropriation by actors.

## 3 CSAO: Profession learning approach

Based on corporate memory life cycle proposed in [6], we propose a profession memory life cycle (Figure 2. ) that considers needs identification, knowledge formalization and dissemination, profession memory use and updating.

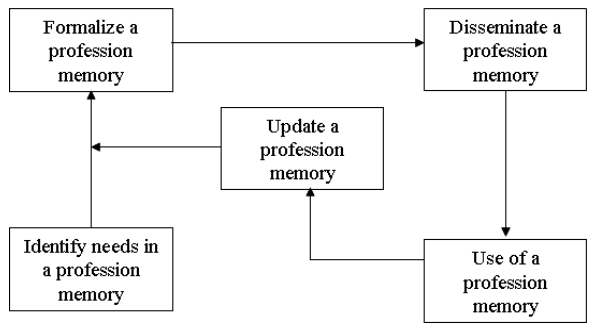

**Figure 2.** Profession memory life-cycle [4].

## 3.1 Identify needs

We suggest localizing knowledge to be formalized, sources and carriers of this knowledge. It is also necessary to find the future users of knowledge and the various types of users if there is. The main activity in this phase is the definition of a knowledge map. According to Aubertin [2] knowledge map allows supplying a structure of cognitive resources of the organization. Three approaches can be used to organize these resources: procedural classification (according to company processes), functional classification (pressed usually on an organization chart), and abstract classification or by domains (information organized about matters, objects and purposes).

Knowledge map design forces the experts involved to sharing the knowledge that they possess, and dissemination of this map of knowledge allows to know the type of knowledge produced in companies and knowledge workers.

## 3.2 Formalize

To formalize a profession memory, we propose a life-cycle composed by [5]:
- Knowledge elicitation: knowledge management responsible team must capture the knowledge, directly by interviewing experts and from organization documents, using knowledge engineering techniques.
- Conceptual model building: according to American Institutes for Research [1], "an application's conceptual model is the "mental map" that the designers build into it in order to present information logically". In our case expertise conceptual model is represented as: activity process, interactions in every stage of the activity process, and problems solving strategy (Figure 3. ). Concepts and constraints models are also defined. They show links between problem solving and domain knowledge.

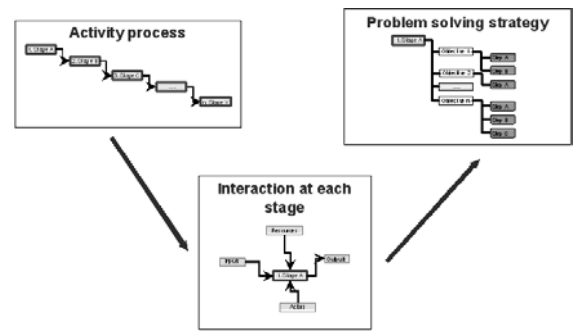

**Figure 3.** Conceptual model.

- Learning model implementation: we propose the construction of a competences tree, based on the same principles of the accomplishment of organization knowledge map. Knowledge management responsible team has to assign to every stage of the process, the group of the competences necessary. Every competence must have assigned a value, which will be compared with the values of the competences of an employee. The profile of an employee and the educational activities had also to be defined.

Accomplishment of each one of these stages, asks strong interactions between experts and organization knowledge responsible team, and involves a big work of knowledge sharing. We propose the creation of a community of practice of experts in the organization. "A community of practice exists because it produces a shared practice as members engage in a collective process of learning" [13].

### 3.3 Update

Communities of practice, will serve to support alive the profession memory and are going to allow updating it. According to Wenger, "communities of practice are important to the functioning of any organization, but they become crucial to those that recognize knowledge as a key asset. From this perspective, an effective organization contains a constellation of interconnected communities of practice, each of them dealing with specific aspects of the company's competency–from the peculiarities of a long-standing client, to manufacturing safety, to esoteric technical inventions. Knowledge is created, shared, organized, revised, and passed on within and among these communities. In a deep sense, it is by these communities that knowledge is "owned" in practice" [13].

### 3.4 Disseminate

Learning system will be composed by an expert module, a student module, a pedagogical model and an interface module. The profession memory will be the expert module and an individual memory, representing the employee competences, will be the student module. The profession memory (expert module) have to be integrated on the employee workspace. Index and learning guides give access to knowledge at any time.

### 3.5 Use

One of our postulates is that the employee has to use alone the learning system. The employee will have no tutor or councilors to guide him during this learning. Our learning system, auto-training, has to have the necessary and sufficient information to allow the employee to realize the training without to consult an expert [5]. The same employee will be who decide if he has already realized successfully the educational activities and will realize an auto-evaluation of the obtained answer (Figure 4. ). Operational learning system, based on employee competences, will decide if he can pass to other stage of activity process.

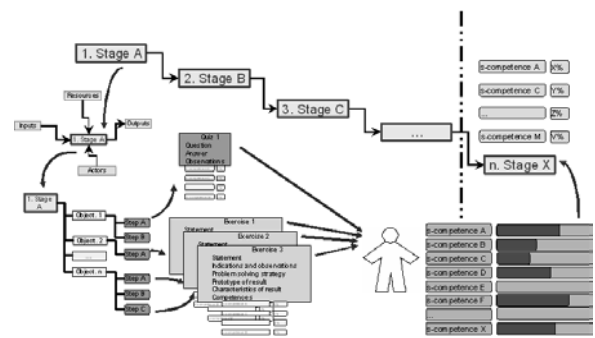

**Figure 4.** Progression process illustration of knowledge learning [5].

## 4 Applications

Textile companies have to deal with a large competition in Textile confection. They have to develop specific expertise. The French Textile Institute (IFTH) helps Textile industry to develop these expertises. It studies organizational learning techniques to help companies. One of their studies is to enhance learning from profession memory. We follow the profession memory life-cycle defined in CSAO for this aim:

1. Experts have been identified. They work on integral and cylindrical knitting without seam. Users are machine workmen in a number of Textile Industries. Cylindrical machines can be bought for that aim. But that supposes changing of a number of current machines and workspace organization in industry. Workmen will not be able also to use machine to do normal knitting. So French textile Institute develops expertise on cylindrical (ot Integral) knitting in normal (no cylindrical) machine.

2. A capitalization of knowledge process with MASK method is applied and knowledge is represented as: process, problem solving, constraints and concepts models. Graphical representation is used.

3. Knowledge so represented is then transformed in training schemas: Learners are conducted to follow a course based on the experts' process of knitting. For each step: they have to solve a problem and follow the main steps of experts' strategy (Figure 5. ).

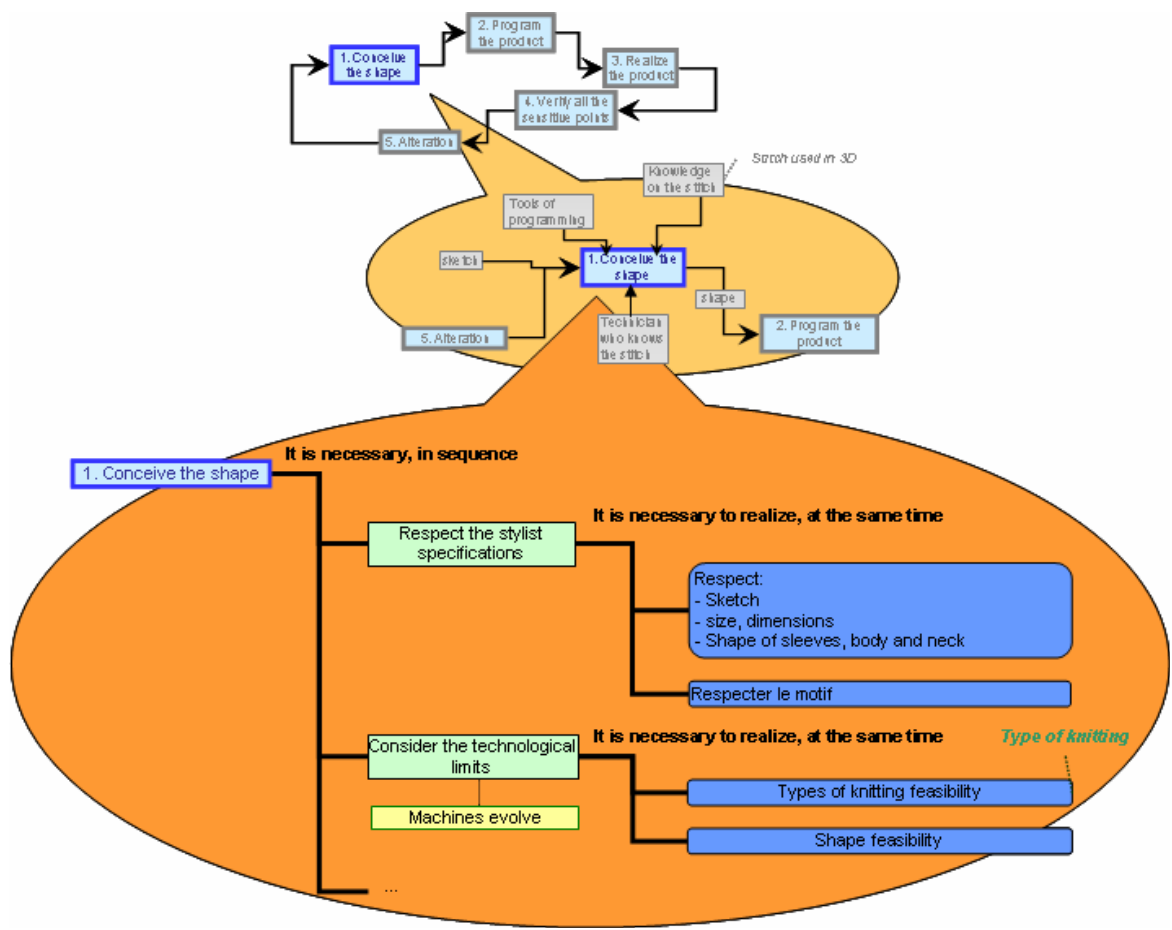

**Figure 5.**   Training schema for Integral knitting learning.

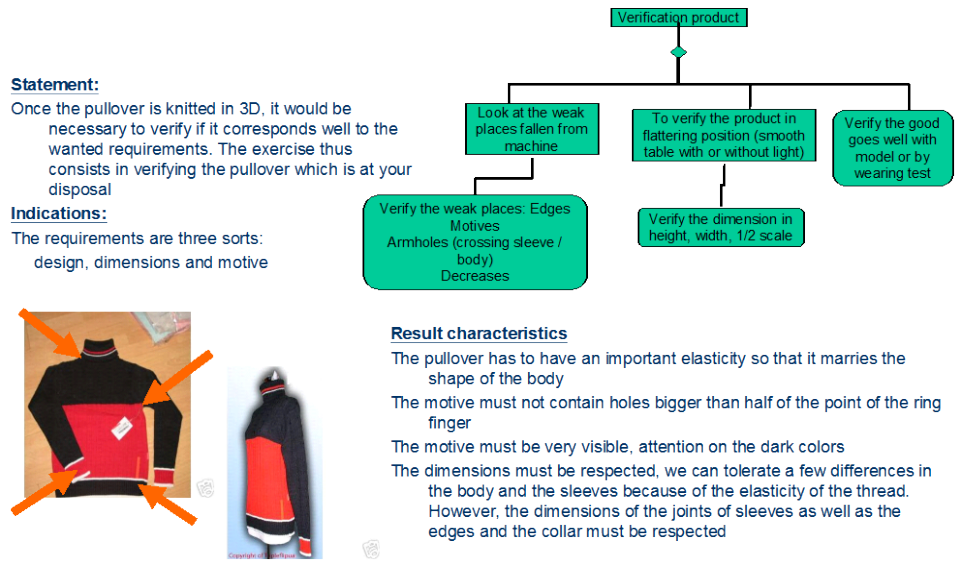

**Figure 6.** Exercise on Integral Knitting

Exercises are also defined for training. A number of indications (as expected results characteristics, constraints to consider, concepts to use etc.) help learners in their training and their auto-evaluation (Figure 6. ).

4. Machine workmen have modified their work, based on their learning. Cylindrical knitting starts to be competitive in their Industry. The knowledge capitalized is represented also as a knowledge book and workmen can consult it as they need.

The French Textile Industry aim at developing several training in different domains based on experts knowledge.

## 5 Feedback

A number of points can be noted from the analysis of the applications of the CSAO approach in knitting:
- ➢ The externalization of knowledge as profession memory helped experts to formalize their know-how and to structure it in order to emphasize the main point to consider in an activity: difficulties, elements to avoid, advices, etc.
- ➢ Experts are very happy to participate in the reorganization of their knowledge in training structure. They need to know the concrete finality of the use of the corporate memory. They are also the only that know how to reorganize the knowledge because they know deeply their domain and their activity.
- ➢ Training sessions have been organized by the IFTH in Textile companies. IFTH Teachers are very happy to have a reference documents that show the main points and the objectives of success knitting. Learner progress easily knowing the goals of steps in knitting and difficulties as expressed in the profession memory.

To summarize our analysis, we can note that training using profession memory promote practical learning better than traditional training that takes more time to enable learner to build optimal strategy to deal with problems of his domain. The CSAO approach we use has been accepted and appreciated by even experts and learners because that guides know-how formalization and transmission.

CSAO approach must be developed in order to enhance self-learning for instance: as base of activity simulations, or to enhance organizational learning (activity communication and practice sharing).

# 6 CONCLUSION

According to Gurteen, "the most effective way to create a knowledge sharing culture – is first to start to practice. Firstly, change the culture. Secondly, put in place the knowledge sharing technology and train and educate people in its effective use. People with the appropriate knowledge sharing mindset and the appropriate knowledge sharing technology to support them will rapidly bring about a knowledge sharing culture that helps organization to better meet its business objectives" [8]. Our work proposes the accomplishment of diverse activities, which must be an example in order that most of organization's members learns to share knowledge and to use the knowledge that the experts want to share.